\pdfoutput=1

\documentclass[11pt]{article}

\usepackage[]{EMNLP2022}

\usepackage{times}
\usepackage{latexsym}

\usepackage[T1]{fontenc}

\usepackage[utf8]{inputenc}

\usepackage{microtype}
\usepackage{graphicx}
\usepackage{amssymb}
\usepackage{amsmath}

\usepackage[normalem]{ulem}
\useunder{\uline}{\ul}{}

\usepackage{inconsolata}

\title{Hate-CLIPper: Multimodal Hateful Meme Classification \\ based on Cross-modal Interaction of CLIP Features}

\author{Gokul Karthik Kumar \quad Karthik Nandakumar \\
  Mohamed Bin Zayed University of Artificial Intelligence (MBZUAI) \\
  Abu Dhabi, UAE \\
  \tt{\{gokul.kumar, karthik.nandakumar\}@mbzuai.ac.ae} \\
  }

\begin{document}
\maketitle
\begin{abstract}
Hateful memes are a growing menace on social media. While the image and its corresponding text in a meme are related, they do not necessarily convey the same meaning when viewed individually. Hence, detecting hateful memes requires careful consideration of both visual and textual information. Multimodal pre-training can be beneficial for this task because it effectively captures the relationship between the image and the text by representing them in a similar feature space. Furthermore, it is essential to model the interactions between the image and text features through intermediate fusion. Most existing methods either employ multimodal pre-training or intermediate fusion, but not both. In this work, we propose the Hate-CLIPper architecture, which explicitly models the cross-modal interactions between the image and text representations obtained using Contrastive Language-Image Pre-training (CLIP) encoders via a \emph{feature interaction matrix} (FIM). A simple classifier based on the FIM representation is able to achieve state-of-the-art performance on the Hateful Memes Challenge (HMC) dataset with an AUROC of 85.8, which even surpasses the human performance of 82.65. Experiments on other meme datasets such as Propaganda Memes and TamilMemes also demonstrate the generalizability of the proposed approach. Finally, we analyze the interpretability of the FIM representation and show that cross-modal interactions can indeed facilitate the learning of meaningful concepts. The code for this work is available at \url{https://github.com/gokulkarthik/hateclipper}.

\end{abstract}

\section{Introduction}

Multimodal memes, which can be narrowly defined as images overlaid with text that spread from person to person, are a popular form of communication on social media \citep{kiela2020hateful}. While most Internet memes are harmless (and often humorous), some of them can represent hate speech. Given the scale of the Internet, it is impossible to manually detect such hateful memes and stop their spread. However, automated hateful meme detection is also challenging due to the multimodal nature of the problem. 

\begin{figure}[t]
\centering
   \includegraphics[width=\linewidth]{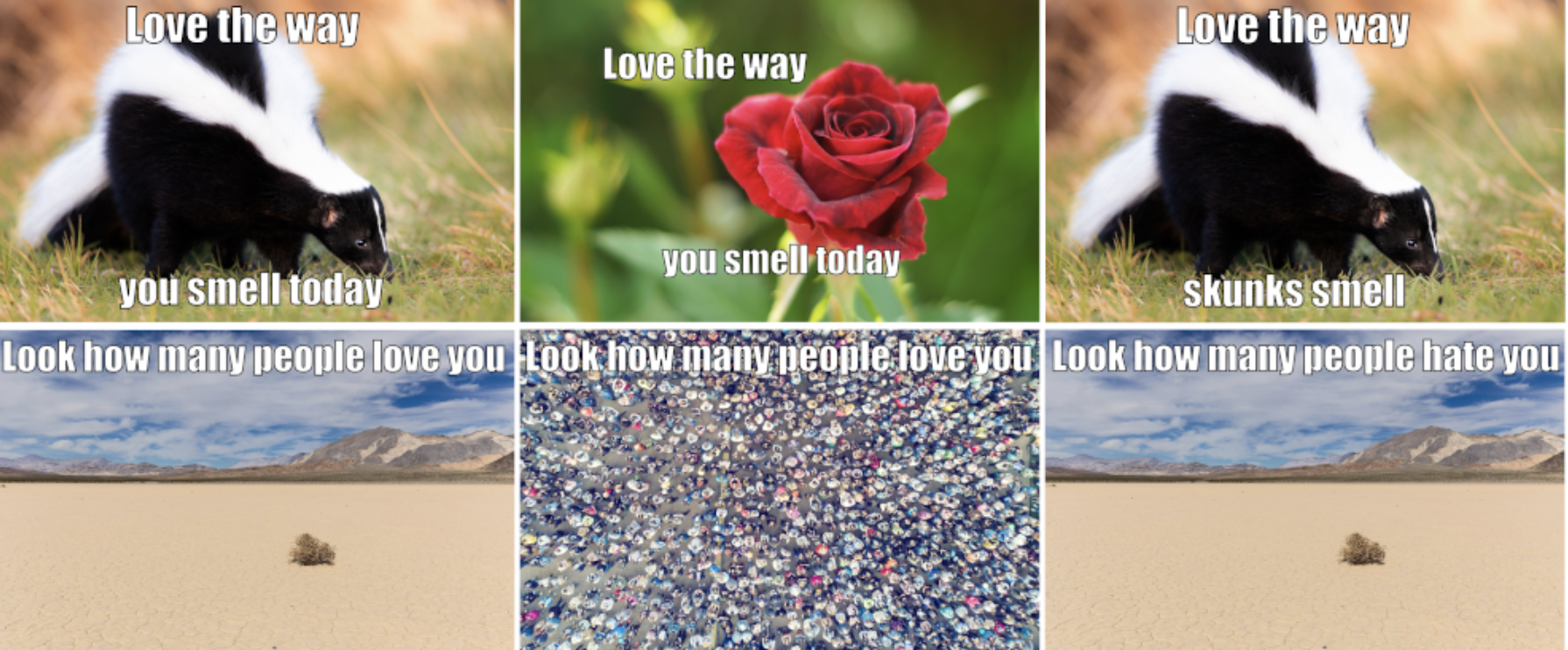}
   \caption{Illustrative (not real) examples of multimodal hateful memes from \citet{kiela2020hateful}. While the memes on the left column are hateful, the ones in the middle are non-hateful image confounders, and those on the right are non-hateful text confounders.}
   \label{fig:example}
\end{figure}

Research on automated hateful meme detection has been recently spurred by the Hateful Memes Challenge competition \citep{kiela2020hateful} held at NeurIPS 2020 with a focus on identifying multimodal hateful memes. The memes in this challenge were curated in such a way that only a combination of visual and textual information could succeed. This was achieved by creating non-hateful ``confounder'' memes by changing only the image or text in the hateful memes, as shown in Figure \ref{fig:example}. In these examples, an image/text can be harmless or hateful depending on subtle contextual information contained in the other modality. Thus, multimodal (image and text) machine learning (ML) models are a prerequisite to achieve robust and accurate detection of such hateful memes.

In a multimodal system, the fusion of different modalities can occur at various levels. In early fusion schemes \cite{kiela2019supervised, lu2019vilbert,li2019visualbert}, the raw inputs (e.g., image and text) are combined and a joint representation of both modalities is learned. In contrast, late fusion approaches \cite{kiela2020hateful}, learn end-to-end models for each modality and combine their outputs. However, both these approaches are not appropriate for hateful memes because the text in a meme does not play the role of an image caption. Early fusion schemes are designed for tasks such as captioning and visual question answering, where there is a strong underlying assumption that the associated text describes the contents of the image. Hateful memes violate this assumption because the text and image may imply different things. We believe that this phenomenon makes the early fusion schemes non-optimal for hateful meme classification. In the example shown in the first row of Figure \ref{fig:example}, the left meme is hateful because of the interaction between the image feature "skunk" and the text feature "you" in the context of the text feature "smell". On the other hand, the middle meme is non-hateful as "skunk" got replaced by "rose" and the right meme is also non-hateful because "you" got replaced by "skunk". Thus, the image and text features are related via common attribute(s). Since modeling such relationships is easier in the feature space, an intermediate fusion of image and text features is more suitable for hateful meme classification.  

The ability to model relationships in the feature space also depends on the nature of the extracted image and text features. Existing intermediate fusion methods such as ConcatBERT \citep{kiela2020hateful} pretrain the image and text encoders independently in a unimodal fashion. This could result in the divergent image and text feature spaces, making it difficult to learn any relationship between them. Thus, there is a need to ``align'' the image and text features through multimodal pretraining. Moreover, hateful meme detection requires faithful characterization of interactions between fine-grained image and text attributes. Towards achieving this goal, we make the following contributions in this paper:

\begin{itemize}
\item We propose an architecture called Hate-CLIPper for multimodal hateful meme classification, which relies on an intermediate fusion of aligned image and text representations obtained using the multimodally pretrained Contrastive Language-Image Pretraining (CLIP) encoders \citep{radford2021learning}. 

\item We utilize bilinear pooling (outer product) for the intermediate fusion of the image and text features in Hate-CLIPper. We refer to this representation as feature interaction matrix (FIM) which explicitly models the correlations between the dimensions of the image and text feature spaces. Due to the expressiveness of the FIM representation from the robust CLIP encoders, we show that a simple classifier with few training epochs is sufficient to achieve state-of-the-art performance for hateful meme classification on three benchmark datasets without any additional input features like object bounding boxes, face detection and text attributes.

\item We demonstrate the interpretability of FIM by identifying salient locations in the FIM that trigger the classification decision and clustering the resulting trigger vectors. Results indicate that FIM indeed facilitates the learning of meaningful concepts.  
\end{itemize}


\begin{figure*}[t]
\centering
   \includegraphics[width=\linewidth]{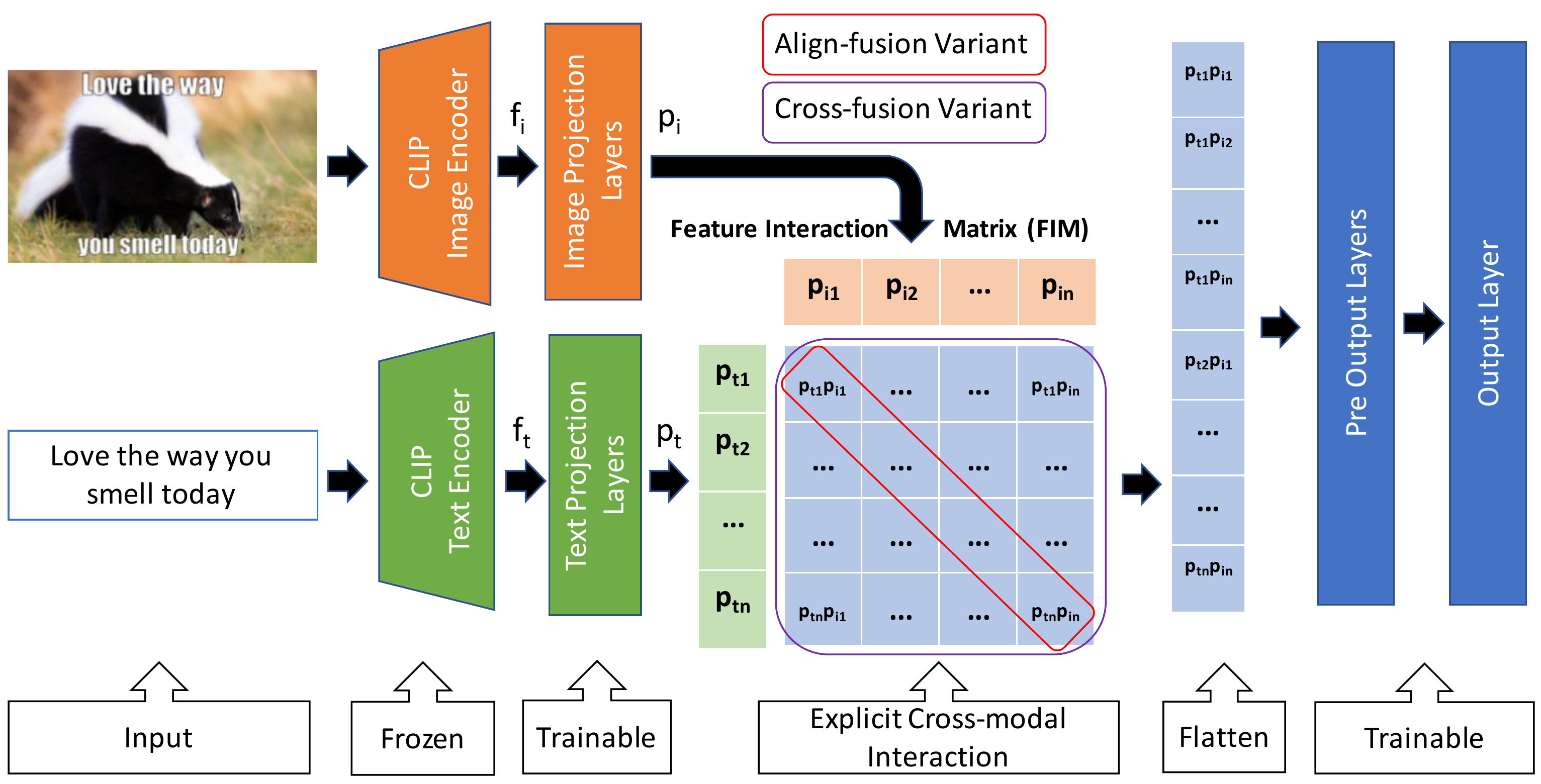}
   \caption{Proposed architecture of Hate-CLIPper for Multimodal Hateful Meme Classification.}
   \label{fig:hate-clip}
\end{figure*}

\section{Related Work}
\label{sec:rel}

The Hateful Memes Challenge (HMC) competition \citep{kiela2020hateful} established a benchmark dataset for hateful meme detection and evaluated the performance of humans as well as unimodal and multimodal ML models. The unimodal models in the HMC competition include: \textbf{Image-Grid}, based on ResNet-152 \citep{he2016deep} features; \textbf{Image-Region}, based on Faster RCNN \citep{ren2017faster} features; and \textbf{Text-BERT}, based on the original BERT \cite{devlin2018bert} features. The multimodal models include: \textbf{Concat BERT}, which uses a multilayer perceptron classifier based on the concatenated ResNet-152 (image) and the original BERT (text) features; \textbf{MMBT} \citep{kiela2019supervised} models, with Image-Grid and Image-Region features; \textbf{ViLBERT} \citep{lu2019vilbert}; and \textbf{Visual BERT} \citep{li2019visualbert}. A late fusion approach based on the mean of Image-Region and Text-BERT output scores was also considered. All the above models were benchmarked on the ``test seen'' split  based on the area under the receiver operating characteristic curve (AUROC) \citep{BRADLEY19971145} metric. The results indicate a large performance gap between humans (AUROC of 82.65 \footnote{https://ai.facebook.com/blog/hateful-memes-challenge-and-data-set/}) and the best baseline using Visual BERT (AUROC of 75.44).

The challenge report \citep{pmlr-v133-kiela21a}, which was released after the end of the competition, showed that all the top five submissions \citep{zhu2020enhance, muennighoff2020vilio, velioglu2020detecting, lippe2020multimodal, s2020detecting} achieve better AUROC than the baseline methods. This improvement was achieved primarily through the use of ensemble models and/or external data and additional input features. For example, \citet{zhu2020enhance} used a diverse ensemble of VL-BERT \citep{su2019vlbert}, UNITER-ITM \citep{chen2019uniter}, VILLA-ITM \citep{gan2020largescale} and ERNIE-Vil \citep{yu2020ernievil} with additional information about entity, race, and gender extracted using Cloud APIs and other models. This method achieved the best AUROC of 84.50 on the ``test unseen'' split. 

\citet{mathias-etal-2021-findings} extended the HMC dataset with  fine-grained labels for protected category  and attack type. Protected category labels include race, disability, religion, nationality, sex, and empty protected category. Attack types were labeled as contempt, mocking, inferiority, slur, exclusion, dehumanizing, inciting violence, and empty attack. \citet{zia-etal-2021-racist} used CLIP \citep{radford2021learning} encoders to obtain image and text features, which were simply concatenated and passed to a logistic regression classifier. Separate classification models were learned for the two multilabel classification tasks - protected categories and attack types. MOMENTA \citep{pramanick2021momenta} also uses representations generated from CLIP encoders, but augments them with the additional feature representations of objects and faces using VGG-19 \cite{simonyan2014very} and text attributes using DistilBERT \cite{sanh2019distilbert}. Furthermore, MOMENTA uses cross-modality attention fusion (CMAF), which concatenates text and image features (weighted by their respective attention scores) and learns a cross-modal weight matrix to further modulate the concatenated features. MOMENTA reports performance only on the HarMeme dataset \cite{s2020detecting}.

Although bilinear pooling \cite{tenenbaum2000separating} (outer product) of different feature spaces has shown improvements for different multimodal tasks \cite{fukui2016multimodal, arevalo2017gated, kiela2018efficient}, it is not well experimented with multimodally pretrained (aligned feature space) encoders like CLIP or for the Hateful Meme Classification task.


\begin{table*}[t]
\centering
\begin{tabular}{|c|c|c|c|c|c|c|}
\hline
\textbf{\# proj. layers} &
  \textbf{\# p.o. layers} &
  \textbf{Fusion} &
  \textbf{Model} &
  \textbf{Dev seen} &
  \textbf{Test seen} &
  \textbf{\# t.params.} \\
\hline
1          & 1             & Concat & Baseline                                   & 76.72          & 79.87  &3.9M        \\
1          & 3             & Concat & Baseline                                   & 79.02          & 83.73   &6M       \\
1          & 5             & Concat & Baseline                                   & 78.6           & 83.8       &8.1M    \\
1          & 7             & Concat & Baseline                                   & 78.63          & 83.29        &10.2M  \\
\hline
1          & 1             & CMAF & MOMENTA                                   & 77.36         & 80.15   &4.5M      \\
1          & 3             & CMAF & MOMENTA                                   & 76.85          & 82     &6.6M     \\
1          & 5             & CMAF & MOMENTA                                   & 79.51           & 83.35     &8.7M      \\
1          & 7             & CMAF & MOMENTA                                   & 78.88          & 82.4        &10.8M \\
\hline
1          & 1             & Cross  & HateCLIPper                                & \textbf{82.62} & 85.12          &1.1B\\
1          & 3             & Cross  & HateCLIPper                                &82.19         & 82.66          &1.1B\\
\hline
1          & 1             & Align  & HateCLIPper                                & 81.18          & 85.46          &2.9M\\
1          & 3             & Align  & HateCLIPper                                & 81.55         & \textbf{85.8}    &5M       \\
1          & 5             & Align  & HateCLIPper                                & 80.88          & 85.46 &7.1M \\
1          & 7             & Align  & HateCLIPper                                & 81.09          & 84.88     &9.2M    \\
\hline
\end{tabular}
\caption{AUROC of Hate-CLIPper variants and other fusion approaches on HMC dataset. Expansions: proj. -> projection; p.o. -> pre-output; t.params. -> trainable parameters; M -> million; B -> Billion.}
\label{tab:result-fb}
\end{table*}

\begin{table*}[!h]
\centering
\begin{tabular}{|c|c|c|c|c|c|c|}
\hline
\textbf{\# proj. layers} &
  \textbf{\# p.o. layers} &
  \textbf{Fusion} &
  \textbf{Model} &
  \textbf{Dev} &
  \textbf{Test} &
  \textbf{\# t.params.} \\
\hline
1          & 1             & Concat & Baseline                                   & 89.9          & 88.93 &4M \\
1          & 3             & Concat & Baseline                                   & 89.9          & 88.82  & 6.1M        \\
1          & 5             & Concat & Baseline                                   & 89.18          & 88.55     &8.2M      \\
1          & 7             & Concat & Baseline                                   & 89.83          & 88.82     &10.3M     \\
\hline
1          & 1             & CMAF & MOMENTA                                   & 89.11         & 88.34   &4.5M      \\
1          & 3             & CMAF & MOMENTA                                   & 89.75          & 88.73   & 6.6M       \\
1          & 5             & CMAF & MOMENTA                                   & 89.11           & 88.34  & 8.7M          \\
1          & 7             & CMAF & MOMENTA                                   & 89.61          & 88.66 &10.8M        \\
\hline
1          & 1             & Cross  & HateCLIPper                                & \textbf{90.98} & \textbf{90.41}   &1.1B      \\
1          & 3             & Cross  & HateCLIPper                                &\textbf{90.98}         & 89.95        &1.1B \\
\hline
1          & 1             & Align  & HateCLIPper                                & 89.11         & 88.34   &2.9M      \\
1          & 3             & Align  & HateCLIPper                                & 89.11         & 88.34        &5M     \\
1          & 5             & Align  & HateCLIPper                                & 89.68          & 88.66 & 7.1M \\
1          & 7             & Align  & HateCLIPper                                & 89.68          & 88.66       &9.2M \\
\hline
\end{tabular}
\caption{Micro F1 scores of Hate-CLIPper variants and other fusion approaches on Propaganda Memes dataset. Expansions: proj. -> projection; p.o. -> pre-output; t.params. -> trainable parameters; M -> million; B -> Billion.}
\label{tab:result-prop}
\end{table*}

\section{Methodology}

Our objective is to develop a simple end-to-end model for hateful meme classification that avoids the need for sophisticated ensemble approaches and any external data or labels. We hypothesize that there is sufficiently rich information available in the CLIP visual and text representations and the missing link is the failure to model the interactions between these feature spaces adequately.  Hence, we propose the Hate-CLIPper architecture as shown in Figure \ref{fig:hate-clip}. In the proposed Hate-CLIPper architecture, the image $i$ and text $t$ are passed through pretrained CLIP image and text encoders (whose weights are frozen after pretraining) to obtain unimodal features $f_i$ and $f_t$, respectively. We use pre-trained CLIP encoders from the original work \cite{radford2021learning}, where the model is trained on Image-Text matching with 400 million <image, text> pairs collected from the Internet.

\noindent \textbf{Trainable Projection Layers}: Note that CLIP
is pre-trained using contrastive learning on 400 million image–text pairs from the Internet. This multimodal pretraining encourages similarity between the feature spaces of the image and its corresponding text caption. However, in the dataset used for pretraining, the image and text pairs usually convey the same meaning, which is not always the case in hateful memes. Therefore, to better model the semantic relationship  between the image and text feature spaces of memes, we further add trainable projection layers at the output of the CLIP image and text encoders. The main purpose of projection layers is not to ensure same dimensionality for both text and image embeddings, but to achieve better alignment between the text and image spaces. While CLIP is already trained to align the two spaces at a high-level, this needs to be further finetuned for the specific task/dataset at hand. Instead of finetuning the entire CLIP model using small datasets, it is more prudent to add projection layers and only learn these projection layers based on the given datasets. These projection layers map the unimodal image and text features $f_i$ and $f_t$ to the corresponding image projection $p_i$ and text projection $p_t$, respectively. The projection layers are designed such that both $p_i$ and $p_t$ have the same dimensionality $n$. The use of customized trainable projection layers after the CLIP encoders is one of the key differences between the proposed architecture and the one used in \cite{zia-etal-2021-racist}.

\noindent \textbf{Modeling Full Cross-modal Interactions}: The important component of the Hate-CLIPper architecture is the explicit modeling of interactions between the projected image and text feature spaces using a \emph{feature interaction matrix} (FIM). The FIM representation $R \in \mathbb{R}^{n \times n}$ is obtained by computing the outer product of $p_i$ and $p_t$, i.e., $R = p_i \otimes p_t$. The FIM can be flattened to get a vector $r$ of length $n^2$ and passed through a learnable neural network classifier to obtain the final classification decision. This approach is different from the traditional concatenation (Concat) technique employed in the literature \cite{zia-etal-2021-racist, pramanick2021momenta}, which simply concatenates the two representations to obtain a vector of length $2n$. Since the FIM representation directly models the correlations between the dimensions of the image and feature spaces, it is better than the Concat approach, where the task of learning these relationships from limited data samples falls on the subsequent classification module. We refer to the fusion of text and image features using the FIM as \emph{cross-fusion}.

\noindent \textbf{Modeling Reduced Cross-modal Interactions}: One of the limitations of the cross-fusion approach is the high dimensionality of the resulting representation, which in turn requires a classifier with a larger number of parameters. The diagonal elements of the FIM $R$ represent the element-wise product between $p_i$ and $p_t$ and has a dimension of only $n$. Note that the sum of these diagonal elements is nothing but the dot product between $p_i$ and $p_t$, which intuitively measures the alignment (angle) between the two vectors. Therefore, a vector representing the diagonal elements of $R$ indicates the alignment between the individual dimensions of $p_i$ and $p_t$, which can still be useful for classification as the encoders that we use are pretrained with the alignment task. Hence, we refer to the fusion of text and image features using only the diagonal elements of FIM as \emph{align-fusion}.

\noindent \textbf{Classification Module}: The output of the intermediate fusion module is a vector $r$ of dimension $d$ (where $d=2n$ for the baseline Concat technique, $d=n^2$ for the cross-fusion approach, and $d=n$ for the align-fusion method). We apply a shallow neural network on this feature vector $r$ to obtain the final output $o$. The shallow neural network consists of a few fully-connected layers (referred to as pre-output layers) and a softmax output layer to produce the final output value $o$. The first layer of this classifier network maps an input $r$ with $d$ dimensions to a common pre-output dimension $m$ and the rest of the pre-output layers have the same number of hidden nodes $m$. Each fully-connected layer is followed by ReLU activation and trained with dropout. For binary classification (hateful vs. non-hateful memes), we optimize the trainable (projection and pre-output) layers by minimizing the binary cross-entropy loss between the output $o$ and the true label $l$. For fine-grained classification (protected category, attack type), we simply add auxiliary output layers and train the model using the total loss for all the classification tasks. 

\section{Experimental Results}
\label{sec:exp}

\subsection{Datasets}
The primary dataset used in our evaluation is the HMC dataset \cite{kiela2020hateful}, which contains 8500 memes in the training set, 500 memes in the development seen split, 540 memes in the development unseen split, 1000 memes in the test seen split, and 2000 memes in the test unseen split. We also evaluate the proposed approach on the Propaganda Memes dataset \citet{sharma2022detecting}, which is a multi-label multimodal dataset with 22 propaganda classes. Finally, to evaluate the multilingual generalizability, we test the performance on TamilMemes \cite{suryawanshi-etal-2020-dataset}, which is a dataset for troll/non-troll classification of memes in the Tamil language. Similar to the HMC dataset, the TamilMemes dataset has meme images and corresponding meme texts that are transliterated from Tamil to English. However, unlike the HMC dataset, the TamilMemes dataset is not compiled with the motive of making only multimodal information useful for target classification.  

\subsection{Setup}
We train Hate-CLIPper and other baselines (concat fusion and attention-based CMAF) based on the train split and evaluate them on the dev-seen and test-seen splits of the HMC dataset using AUROC as the evaluation metric. For the Propaganda Memes and the Tamil Memes dataset, the micro F1 score is used as the evaluation metric to ensure a fair comparison with results reported in the literature. We use TorchMetrics library\footnote{\url{https://torchmetrics.readthedocs.io}} to compute all the evaluation metrics. For multi-label classification in Propaganda Memes dataset, we set `mdmc\_average' to `global' in computing the micro-F1 score, which does the global average for multi-dimensional multi-class inputs. We use Pytorch on NVIDIA Tesla A100 GPU with 40 GB dedicated memory and CUDA-11.1 installed. The hyper-parameter values for all models are shown in Table \ref{tab:hyper}, which are chosen based on the manual tuning with respect to the target evaluation metric of the validation set. We use ViT-Large-Patch14 based CLIP model consistently for all the experiments in Tables  \ref{tab:result-fb} \& \ref{tab:result-prop}. The models experimented in Table \ref{tab:result-fb} took around 30 minutes (median is 30 minutes and longest is 32 minutes) for the combined training and evaluation. To do a fair evaluation, we use the same evaluation metric as in the previous works for the corresponding datasets.


\begin{table}[!h]
\begin{center}
\begin{tabular}{c c} 
\hline
Hyperparameter        & Value \\
\hline
Image size & 224 \\
Pretrained CLIP model & ViT-Large-Patch14  \\
Projection dimension ($n$) & 1024 \\
Pre-output dimension ($m$) & 1024 \\
Optimizer & AdamW \\
Maximum epochs & 20 \\
Batch size & 64 \\
Learning rate & 0.0001 \\
Weight decay & 0.0001 \\
Gradient clip value & 0.1 \\
\hline
\end{tabular}
\end{center}
\caption{Hyperparameter configuration for HateCLIPer and other baselines.}
\label{tab:hyper}
\end{table}

\subsection{Key Findings}

\begin{table}[h]
\centering
\begin{tabular}{|c|c|c|}
\hline
  \textbf{Model} &
  \textbf{Dev Seen} &
  \textbf{Test Seen} \\
\hline
Human                                      & -              & 82.65          \\
Image-Grid                                 & 52.33          & 53.71          \\
Image-Region                               & 57.24          & 57.74          \\
Text-BERT                                  & 65.05          & 69             \\
Late Fusion                                & 65.07          & 69.3           \\
Concat BERT                                & 65.88          & 67.77          \\
MMBT-GRID                                  & 66.73          & 69.49          \\
MMBET-Region                               & 72.62          & 73.82          \\
ViLBERT CC                                 & 73.02          & 74.52          \\
Visual BERT COCO                           & 74.14          & 75.44          \\
CLIP-ViT-L/14-336px  & 77.3           & -              \\
SEER-RG-10B           & 73.4           & -              \\
FLAVA w/o init      & 77.45          & -              \\
\hline
\end{tabular}
\caption{AUROC of different models on the HMC dataset, compiled from \citet{kiela2020hateful, goyal2022vision, singh2021flava}.}
\label{tab:result-fb-compiled}
\end{table}

\begin{table}[!h]
\centering
\begin{tabular}{|c|c|c|}
\hline
  \textbf{Model} &
  \textbf{Test} \\
\hline
Random                                     & 7.06          \\
Majority Class                                & 29.04          \\
ResNet-152                              & 29.92         \\
FastText                                  & 33.56           \\
BERT                                & 37.71           \\
FastText + ResNet-152                                & 36.12          \\
BERT + ResNet-152                                  & 38.12         \\
MMBT & 44.23 \\
ViLBERT CC & 46.76 \\
VisualBERT COCO & 48.34 \\
RoBERTa & 48 \\
RoBERTa + embeddings & 58 \\
Ensemble of BERT models & 59 \\
\hline
\end{tabular}
\caption{Micro F1 scores of different models in Propaganda Memes dataset, compiled from \citet{sharma2022detecting, dimitrov2021detecting}}
\label{tab:result-prop-compiled}
\end{table}

When we interpret Tables \ref{tab:result-fb} \& \ref{tab:result-prop} in conjunction with Tables \ref{tab:result-fb-compiled} \& \ref{tab:result-prop-compiled} respectively, we can clearly see that the performance of intermediate fusion with the CLIP encoders is better that that of several early fusion approaches such as MMBT, ViLBERT, and VisualBERT as well as late fusion methods. For instance, on the HMC dataset, the best early fusion approach (Visual BERT) had an AUROC of 75.44 and late fusion method had an AUROC of 69.3 on the test set. These AUROC values are significantly lower than AUROC of the proposed align (intermediate) fusion scheme, which is 85.8. In fact, all the intermediate fusion methods considered in Table \ref{tab:result-fb} clearly outperform the early and late fusion methods reported in Table \ref{tab:result-fb-compiled}. These results strongly support the claim that intermediate fusion is more suitable for hate classification. 

From Tables \ref{tab:result-fb} \& \ref{tab:result-fb-compiled}, it is clear that cross-fusion and align-fusion variants of Hate-CLIPper achieve the best AUROC for both the evaluation sets of HMC dataset, which is also better than the reported human performance. This trend is also consistent when we replaced ViT-Large-Patch14 with ViT-Base-Patch32. Despite having only $n$ multimodal features, align-fusion performs significantly better than concat-fusion with $2n$ multimodal features and is closer to cross-fusion with $n^2$  multimodal features. This signifies the importance of pre-aligned image and text representations of CLIP. Hence, for low computational resource conditions, it would be appropriate to replace cross-fusion with align-fusion in the Hate-CLIPper framework.

Our results also show that a single projection layer for each modality and a shallow neural network (1 or 3 layers) for the classifier is sufficient to achieve good performance. This shows that the discriminative power of Hate-CLIPper is mainly a consequence of modeling the interactions between text and image features from CLIP encoders using cross and align fusion. The results on the Propaganda Memes dataset also confirm the same findings. Although the differences between the various configurations shown in Table \ref{tab:result-prop} are marginal, the performance of the proposed approach is a significant leap compared to those reported in the literature (see Table \ref{tab:result-prop-compiled}).

As noted in Section \ref{sec:rel}, methods proposed in \cite{zia-etal-2021-racist} and \cite{pramanick2021momenta} are the closest to the proposed approach since both of them use CLIP encoders. Results in Table \ref{tab:result-fb} show that under the same experimental setup, the proposed approach is better than the cross-modal attention fusion (CMAF) scheme used in MOMENTA \cite{pramanick2021momenta}, when no additional information is utilized. If additional information is available, our proposed approach can also leverage them in the same way (using intra-modal fusion) as MOMENTA. The work in \cite{zia-etal-2021-racist} claims that train and development seen splits were used for training and development unseen split was used for evaluation. However, a careful analysis of the published code for \cite{zia-etal-2021-racist} indicates that 400 out of 540 memes in development unseen split (74\%) are also included in the development seen split. Since a fair comparison is not possible under these circumstances, we ignore all the results of \citet{zia-etal-2021-racist}.

Our ablation experiments (i) with unfrozen CLIP encoders (AUROC of <63), and (ii) Non-CLIP encoders (mBERT \cite{devlin2018bert}, VIT \cite{dosovitskiy2021an}) (AUROC of <59) resulted significantly poor scores in the HMC dataset.

\begin{figure*}[]
\centering
   \includegraphics[width=1\linewidth]{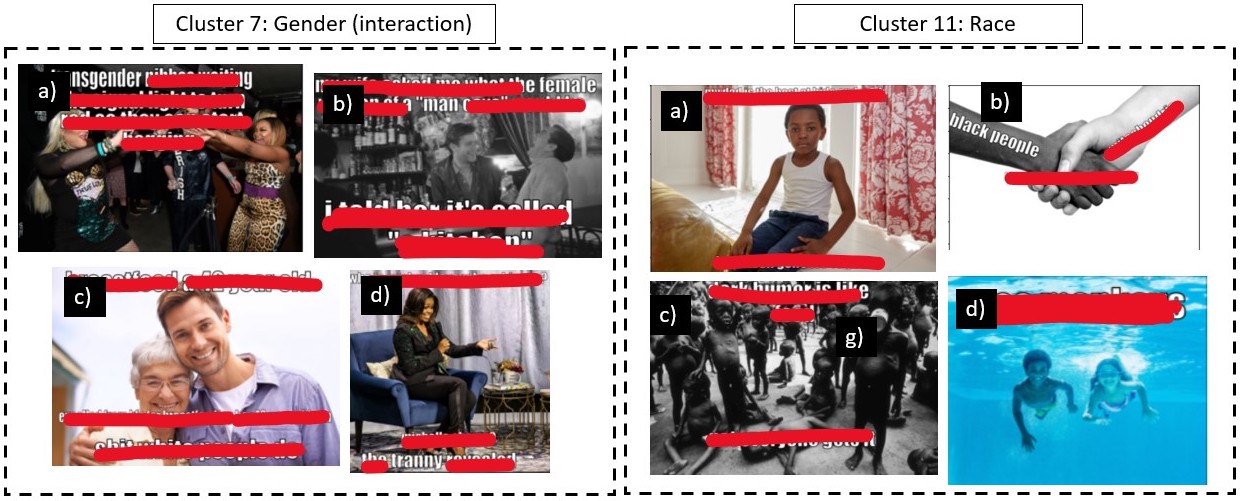}
   \includegraphics[width=1\linewidth]{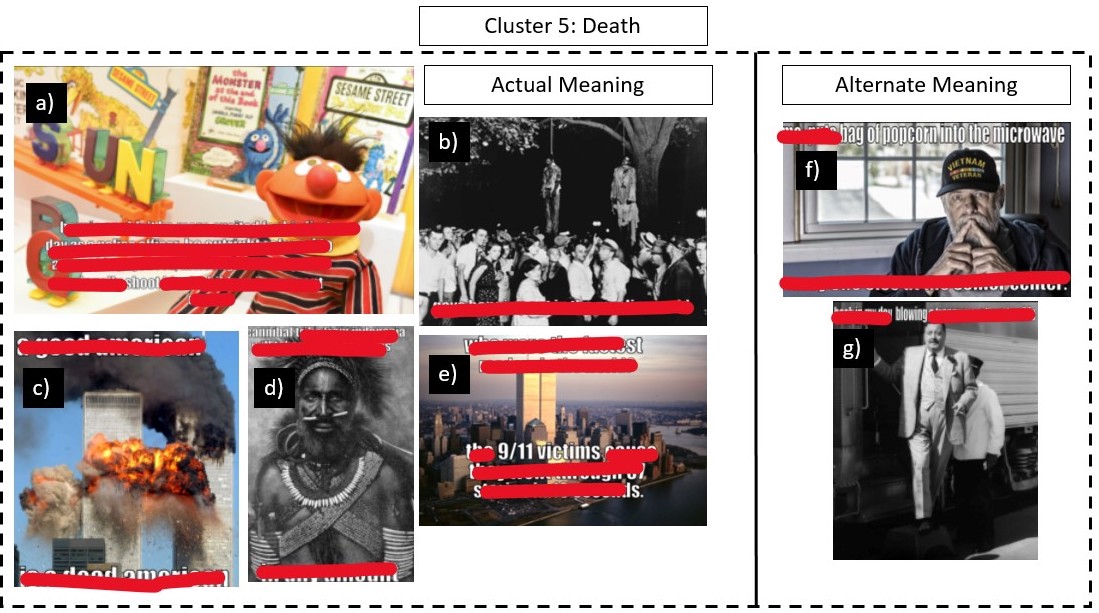}
   \includegraphics[width=1\linewidth]{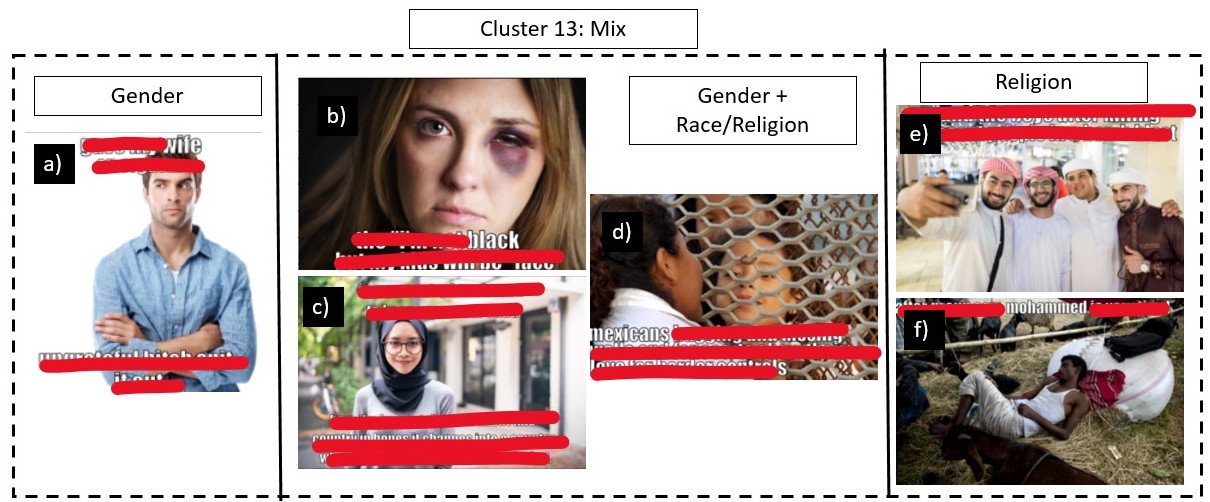}
   \caption{Hateful memes clustered by K-means clustering algorithm (number of clusters = 15) based on the trigger vector of Hate-CLIPper with cross-fusion. Featuring hateful examples with all the original text in this place would be distasteful; hence the meme text is masked with the exception of few words that are required for discussion. However, the reader can choose to look at the non-censored memes in the appendix}
   \label{fig:memes-exp}
\end{figure*}

\subsection{Multilinguality}

The baseline evaluations on TamilMemes dataset  \cite{suryawanshi-etal-2020-dataset} used only image based classifiers such as ResNet \cite{he2016deep} and MobileNet \cite{Howard2017MobileNetsEC} and their test set (300 memes) is different from the released test set (667 memes). Hence, they are not directly comparable to the proposed approach. \cite{Hegde2021UVCEIIITTDravidianLangTechEACL2021TT} proposed a multimodal approach for TamilMemes classification. They used pretrained ViT \cite{dosovitskiy2021an} and BERT \cite{devlin2018bert} as encoders to get the image and text features, respectively. These features were concatenated and used for classification. This model achieved a micro F1 score of 47 on the test set. It is critical to note that the Hate-CLIPper also uses the same encoders of ViT and BERT but they are multimodally pretrained with the CLIP loss. Thus, the Hate-CLIPper achieves state-of-the-art performance with a micro F1 score of 59 on the test set. 

For the TamilMemes dataset, both cross and align fusion had the same performance as concat fusion. This could be due to the fact that the pre-aligned text space of CLIP has never encountered Tamil-to-English transliterated text data. Consequently, the resulting text and image feature spaces are not well-aligned, making it difficult to model the relationship between the two feature spaces. Replacing the CLIP text encoder with multilingual BERT \cite{devlin2018bert} also did not lead to any further performance improvement, which can again be attributed to the feature space misalignment caused by the lack of multimodal pretraining using the image and corresponding Tamil text pairs.

\section{Interpretability}

To determine the interpretability of the feature interaction matrix (FIM), we employ the following simple approach. First, we compute a $n^2$-dimensional binary trigger vector for each hateful meme, where a value of $1$ indicates that the specific element in the FIM $R$ is salient for determining if the given input belongs to the hateful class. These trigger vectors are then clustered into groups using a K-means clustering algorithm. We manually examine these clusters to determine if most samples within a cluster have a common underlying pattern.

To compute the trigger vector, we first reset the feature interaction matrix $R$ to zero values and evaluate the gradient of the loss function for the non-hateful class with respect to $R$. Let $D \in \mathbb{R}^{n \times n}$ denote the model-specific gradient matrix. Each element in the $D$ matrix represents the direction (positive/negative) of the corresponding element in $R$ matrix towards the hateful class. We then binarize the $D$ matrix by setting all elements in the top-20 and bottom 20 percentiles (based on magnitude) to value $1$ and assigning $0$ values to all the other elements. Then, for each hateful meme $(i,t)$ in the training set, we perform one forward pass through the Hate-CLIPper and compute the meme-specific FIM $R$. Again we binarize the $R$ matrix by setting all elements in the top-10 and bottom-10 percentiles (based on magnitude) to value $1$ and assigning $0$ values to all the other elements. Finally, the trigger matrix $T$ is computed for each meme as the element-wise (Hadamard) product of binarized $D$ and $R$ matrices, i.e., $T = D \odot R$. This trigger matrix is then flattened to obtain the trigger vector corresponding to a meme. We apply K-means clustering algorithm available in Scikit-Learn \cite{scikit-learn} on the trigger vectors to group the hateful memes. Samples from the resulting groups of memes are shown in Figure \ref{fig:memes-exp}.

From Figure \ref{fig:memes-exp}, we observe that clusters 5, 7, and 11 contain memes related to the same concept. It is interesting to note that Hate-CLIPper is able to produce similar features for the same concept expressed in different modalities. For example in cluster 5, which is characterized by the concept `death', we can see some memes representing `death' only in images (b) and other memes  representing the same concept only in text (a, d, e). Furthermore, note that the memes (f) and (g), under the same cluster, do not directly relate to death, but the meme texts could hint toward death related events (blow -> blast; popcorn sounds -> bullet sounds) in different contexts. With the clustered memes, we can also identify the positions in FIM $R$, which get activated for the matching concepts. However, some of the clusters are ambiguous. For example, cluster 13 has memes from different concepts. Also, when the clusters have less than 3 memes or greater than 10 memes, they exhibit greater diversity in terms of the underlying concepts and are not useful for the explanation.

\section{Conclusion}
In this work, we emphasized the need for intermediate fusion and multimodal pretraining for hateful meme classification. We proposed a simple end-to-end architecture called Hate-CLIPper using explicit cross-modal CLIP representations, which achieves the state-of-the-art performance quickly in 14 epochs with just 4 trainable layers (1 image projection, 1 text projection, 1 pre-output, and 1 output) in the Hateful Memes Challenge dataset, Moreover, our model does not require any additional input features like object bounding boxes, face detection, text attributes, etc. We also demonstrated similar performance in multi-label classification based on the Propaganda Memes dataset. Finally, we performed preliminary studies to evaluate the interpretability of cross-modal interactions.

\section{Limitations}

From an ethical perspective, the concept of hate speech itself is quite subjective and it is often difficult to draw a clear line between what is hateful and non-hateful. On the technical front, the accuracy of hateful meme classifiers is still far from satisfactory even on carefully curated benchmark datasets, which impedes real-world deployment. Apart from these general limitations, the proposed Hate-CLIPper framework for hateful meme classification also has several specific limitations. Firstly, handling the high dimensionality of the feature interaction matrix is a computational challenge. For $n=1024$ and $m=1024$, this requires a model with a billion parameters ($O(n^2m)$). Fortunately, the align-fusion approach performs quite close to the cross-fusion method and requires only $O(nm)$ parameters. The CLIP encoders used in Hate-CLIPper are well-trained on a massive dataset in English. Such models are rarely available for low-resource languages, limiting their direct applicability for such languages. While the multilingual experiment highlights the issues arising from misaligned text and image feature spaces, more thorough ablation studies are required to understand the ability of learnable projection layer(s) to overcome this misalignment. Furthermore, the proposed approach to judge interpretability is simple and ad-hoc and a more systematic evaluation of explainability is needed. The fine-grained labels \citet{mathias-etal-2021-findings} have not been utilized for  FIM interpretation.
\clearpage

\bibliography{custom}
\bibliographystyle{acl_natbib}

\clearpage

\appendix
\section{Appendix}

\subsection{Variations of HateCLIPper}

We experimented with several training/architectural modifications to the core Hate-CLIPper framework proposed in the main paper:

\begin{enumerate}
    \item \textbf{Pretraining with captions:} We generated captions that describe each image in the Hateful Memes Challenge dataset with the state-of-the-art transformer based image captioning model, OFA \cite{Wang2022UnifyingAT}. Then, we pretrained the image and text encoders of Hate-CLIPper using contrastive loss between the meme images and the generated captions like \citet{radford2021learning}. Then, finetuning on the target dataset using meme images and meme texts is done as usual.
    
    \item \textbf{Finetuning with captions:} We incorporated the generated captions during finetuning of Hate-CLIPper in different ways: (1) replacing image with generated captions and image encoder with the same text encoder, (2) concatenating generated caption with the meme text (3) concatenating features from "meme image + meme text" flow and "meme image+ generated caption" flow.
    
    \item \textbf{Unimodal losses:} Linear output layers, for hateful meme classification, are added on top of the image and text projection layers of Hate-CLIPper and the corresponding unimodal losses are jointly optimized with the original multimodal
     loss as recommended by \citet{Ma2022AreMT}.
    
    \item \textbf{Fine-grained losses:} Linear output layers, for fine-grained hateful meme classification, are added in parallel to the output layer of Hate-CLIPper, and the corresponding fine-grained losses are jointly optimized with the original loss. This is done using fine-grained labels provided by \citet{mathias-etal-2021-findings}.
    
    \item \textbf{Data augmentation:} We identify the text bounding box regions in the meme images using EAST \cite{Zhou2017EASTAE} and replace them with either average pixel value masks or inpainting using Navier-Stokes\footnote{\url{https://docs.opencv.org/4.x/d7/d8b/group__photo__inpaint.html}} based method and finetune the Hate-CLIPper as usual.
\end{enumerate}

Although, the above mentioned variations are backed by some reasoning, they either produced the same results or slightly degraded the performance. 

\begin{figure*}[]
\centering
   \includegraphics[width=\linewidth]{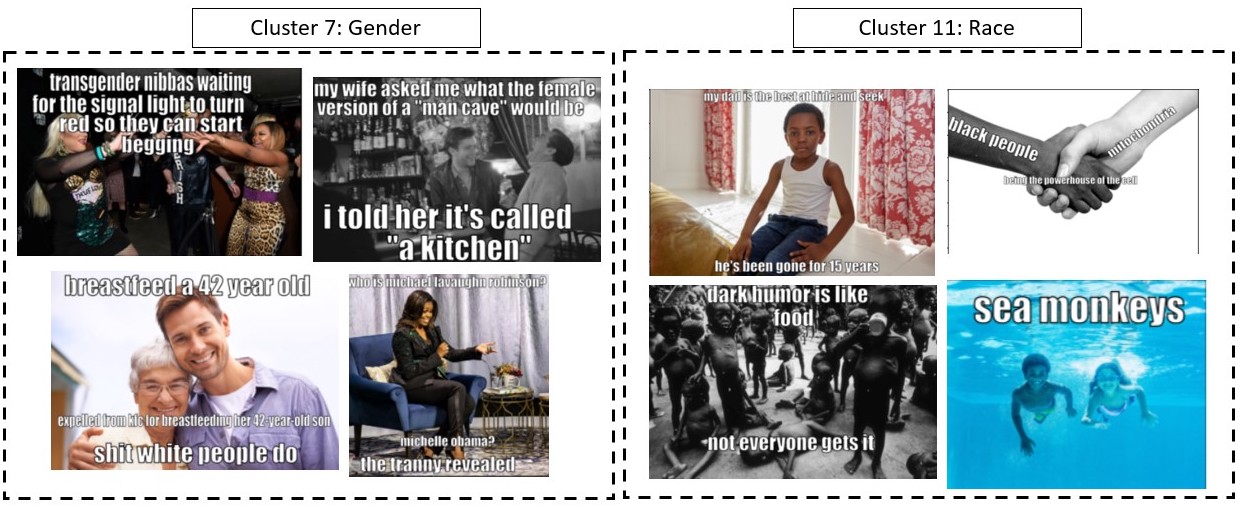}
   \includegraphics[width=\linewidth]{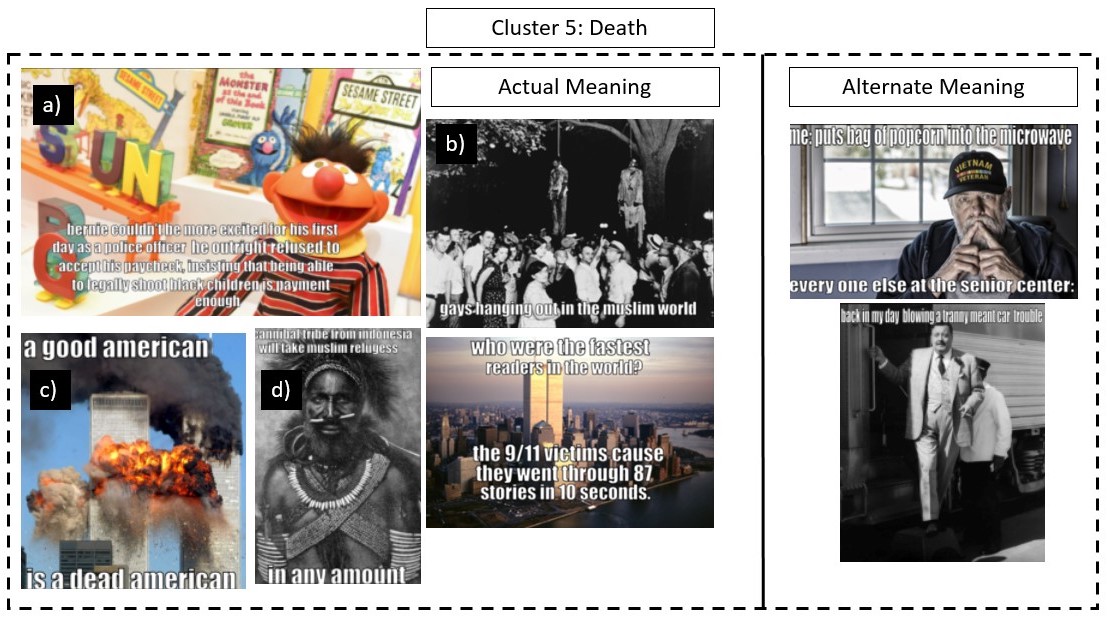}
   \includegraphics[width=\linewidth]{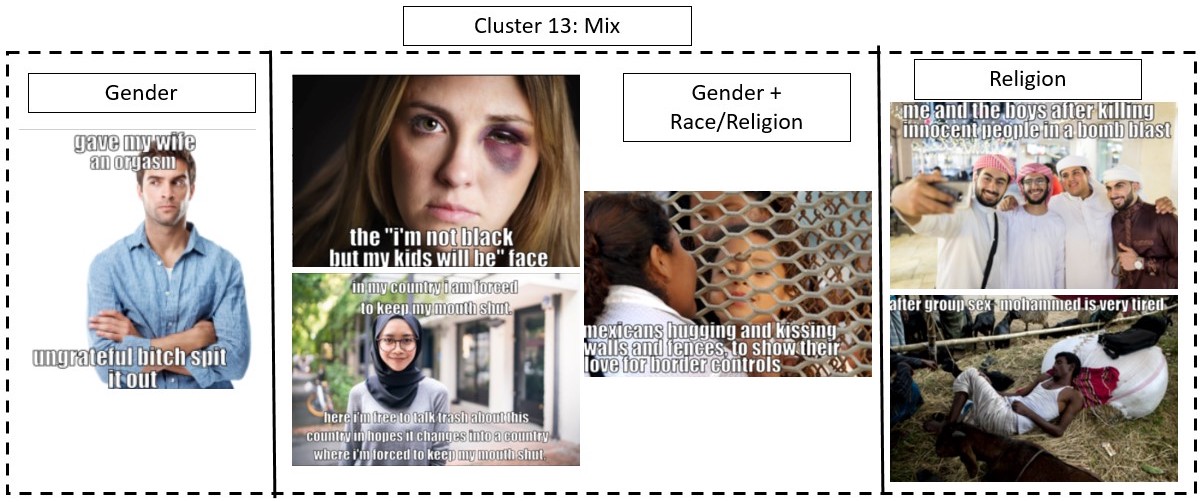}
   \caption{Hateful memes clustered by K-means clustering algorithm (number of clusters = 15) based on the trigger vector of Hate-CLIPper with cross-fusion. This non-censored version is just for more understading and the reader can choose to skip this figure as it features distasteful content.}
   \label{fig:memes-exp}
\end{figure*}



\end{document}